# A data-driven approach to inferring travel trajectory during peak hours in urban rail transit systems

Jie He[1], Yong Qin[2], Jianyuan Guo*[,1], Xuan Sun[3], Xuanchuan Zheng[4]

1 School of Traffic and Transportation, Beijing Jiaotong University, Haidian District, Beijing 100044, China
2 State Key Laboratory of Advanced Rail Autonomous Operation, Beijing Jiaotong University, Haidian District, Beijing 100044, China
3 China Railway Signaling and Communication Research and Design Institute, Fengtai District, Beijing 100070, China
4 Beijing Urban Construction Design & Development Group Co., Xicheng District, Beijing 100032, China
＊ Correspondence: jyguo@bjtu.edu.cn (Jianyuan Guo)

**Abstract**: Refined trajectory inference of urban rail transit is of great significance to the operation organization. In this paper, we develop a fully data-driven approach to inferring individual travel trajectories in urban rail transit systems. It utilizes data from the Automatic Fare Collection (AFC) and Automatic Vehicle Location (AVL) systems to infer key trajectory elements, such as selected train, access/egress time, and transfer time. The approach includes establishing train alternative sets based on spatio-temporal constraints, data-driven adaptive trajectory inference, and trave l trajectory construction. To realize data-driven adaptive trajectory inference, a data-driven parameter estimation method based on KL divergence combined with EM algorithm (KLEM) was proposed. This method eliminates the reliance on external or survey data for parameter fitting, enhancing the robustness and applicability of the model. Furthermore, to overcome the limitations of using synthetic data to validate the result, this paper employs real individual travel trajectory data for verification. The results show that the approach developed in this paper can achieve high-precision passenger trajectory inference, with an accuracy rate of over 90% in urban rail transit travel trajectory inference during peak hours.

# 1 Introduction

Urban rail transit is an important mode of public transportation. The passenger volume of urban rail transit in China exceeded 32 billion last year, accounting for more than 30% of the total urban transportation passenger volume (Transport, 2024). Similarly, urban rail transit is showing high demand worldwide. In 2024, overall demand for the London Underground recovers to 97%, showing a strong relative recovery (London, 2024). Along with the climb in passenger traffic, especially during peak hours, it is also prone to congestion on platforms and trains, leading to the problem of uncertainty in passenger travel. This affects the operational efficiency of the system and puts higher demands on urban rail operation and management.

Uncertainty in boarding options results in the inability to identify the train on which a passenger is traveling. As a result, it is impossible to determine passengers' complete travel chains or accurately estimate the number of passengers at stations or on trains. This travel information is crucial for operational organization. For instance, tracking the impact of high-risk passengers passing through trains and stations during a pandemic (Liu et al., 2020), optimizing train operations and travel guidance to alleviate congestion (Yin et al., 2019), and enhancing passenger comfort and safety (Wang, Wu, Zhao, Peng, & Lin, 2019) all require a thorough understanding of passenger travel trajectories, particularly their train choices.

However, due to the operational characteristics of urban rail transit and the format of internal data records, AFC data only retains information on passengers' entry and exit times and stations. Therefore, it is impossible to directly obtain detailed trajectory information of passengers within the urban rail transit system. Other information collection methods also face a little constraint. For example, although WiFi or Bluetooth-based positioning technologies can track passengers' movement trajectories within stations (Zhao et al., 2022), their accuracy is limited by signal coverage and device penetration rates. On the other hand, methods such as video surveillance (Hsu, Wang, & Perng, 2020) or manual surveys (B. Li, Yao, Yamamoto, Huan, & Liu, 2020) can provide more detailed passenger behavior data, but they are costly and difficult to implement on a large scale. These limitations make it difficult to comprehensively and accurately track the travel trajectories of urban rail passengers. Therefore, overcoming data collection constraints and developing more efficient and rigorous methods for passenger trajectory inference is a significant challenge.

Some existing studies are able to mine individual trajectories using AFC data. For example, some studies infer the selected trains by establishing allocation models (Hamdouch & Lawphongpanich, 2008; Hörcher, Graham, & Anderson, 2017; Kusakabe, Iryo, & Asakura, 2010; Nuzzolo, Crisalli, & Rosati, 2012; Y. Sun &

Schonfeld, 2016; Zhou & Xu, 2012), or combine AVL data with spatiotemporal constraints to estimate travel routes (Luo, Bonnetain, Cats, & van Lint, 2018; L. Sun, Lee, Erath, & Huang, 2012; Y.-S. Zhang & Yao, 2015). Other studies achieve more detailed trajectory mining by incorporating the exploration of passengers' internal movements (Tuncel, Koutsopoulos, & Ma, 2023; Zhu, Koutsopoulos, & Wilson, 2017a, 2017b, 2021). However, existing research still faces the following challenges: (1) Due to the complexity of transfer scenarios, the accuracy of complete trajectory inference needs further improvement. (2) There is a heavy reliance on external or manual survey data, which affects the accuracy of location state information, such as egress time. (3) As the lack of real travel trajectory data, it is difficult to validate the results of individual trajectory inference, making it challenging to verify the accuracy of the inference model.

Therefore, this paper aims to achieve fully data-driven travel trajectory inference during peak hours, and thus construct a complete individual travel chain for passengers. The main contributions of this paper are as follows:

(1) We develop a travel trajectory inference approach for both transfer and non-transfer scenarios, and propose a data-driven parameter estimation method based on KL divergence combined with EM algorithm (KLEM), which solves the problem of dependence on external data or survey data. Thus, a complete individual travel chain can be constructed through fully data-driven trajectory.

(2) To address the challenge of verifying the accuracy of the inference model, we recruit volunteers to obtain real individual travel trajectory data for experimental verification. The experimental results show that the accuracy of trajectory inference of this method reaches more than 90%, which verifies the accuracy and effectiveness of the proposed approach.

The rest of this paper is organized as follows. Section 2 provides a literature review. Section 3 presents the methodology. Section 4 is the case study. Section 5 summarizes the article.

## 2 Literature review

In recent years, with the wide application of Automatic Fare Collection (AFC) data and Automatic Vehicle Location (AVL) data in urban rail transit, there has been an increasing number of related researches on passenger travel aspects. One of the important topics is individual travel trajectory inference research.

For the purpose of travel trajectory inference, it can be mainly divided into travel pattern inference, destination inference, route inference, and train selection inference.

Travel pattern mainly explores the tools and modes chosen by individuals (Cats & Ferranti, 2022; Ma, Wu, Wang, Chen, & Liu, 2013; Weng, Liu, Song, Yao, & Zhang, 2018; Zhao, Qu, Zhang, Xu, & Liu, 2017). Some studies have utilized unsupervised methods to classify passengers through smart card data to infer travel patterns (Cats & Ferranti, 2022; Zhao et al., 2017). The main focus is on extracting features from the input data to realize individual travel pattern recognition. Then it provides the relevant operators with adjustment strategies regarding route planning and so on.

In terms of destination inference, some studies use Markov models and related extension methods to predict the next stage of travel based on history information (Asahara, Maruyama, Sato, & Seto, 2011; Gambs, Killijian, & del Prado Cortez, 2012; Mo, Zhao, Koutsopoulos, & Zhao, 2021), or to predict the sequence of individual travel (Han & Sohn, 2016). These studies often apply Markov models and their variants to forecast an individual's next travel segment. Such research facilitates personalized travel recommendation applications.

In terms of route choice, some studies focus on multi-path inference between OD pairs. These studies often use Logit model and related improvements for route inference (Raveau, Muñoz, & De Grange, 2011; Shi, Pan, He, & Liu, 2023; Si, Zhong, Liu, Gao, & Wu, 2013; Su, Si, Zhao, & Li, 2022). They typically consider factors influencing travel route choices, such as travel time, distance, and transfer frequency, and then calculate the utility of different paths. Other studies integrate Bayesian frameworks with Monte Carlo method to infer route choices by calculating the probability of each path (Kapatsila, Bahamonde-Birke, van Lierop, & Grisé, 2025; L. Sun, Lu, Jin, Lee, & Axhausen, 2015; Tian, Zhu, & Song, 2024; Y.-S. Zhang & Yao, 2015). Additionally, some research applies unsupervised clustering (Chen, Cheng, Jin, Trépanier, & Sun, 2023) or fuzzy matching method (Wu et al., 2019) to allocate passenger flow to routes. However, these approaches still cannot obtain precise and complete individual travel trajectory. Furthermore, these approaches are relatively dependent on assumptions and external data.

Some studies also incorporate external location data (El-Tawab, Oram, Garcia, Johns, & Park, 2016; Gu et al., 2021; Lesani & Miranda-Moreno, 2018; Myrvoll, Håkegård, Matsui, & Septier, 2017; Zhao et al., 2022). Some use WiFi data to predict the number of passengers inside public transportation carriages (Myrvoll et al., 2017), while others integrate WiFi data and network topology to infer spatiotemporal trajectories (Gu et al., 2021), or use trajectory similarity calculations to identify passenger travel paths (Zhao et al., 2022). However, these methods face issues related to external data reliability. Additionally, within urban rail transit, signals and positioning systems often struggle to accurately reflect individual locations, and crowding can lead to data obstruction and high noise levels.

In terms of train inference, the focus is primarily on inferring the train chosen by passengers.

In the early stages, some studies estimated train selected using allocation models (Hamdouch & Lawphongpanich, 2008; Nuzzolo et al., 2012). Due to the complexity of the problem, certain simplifications were made, such as assuming a non-transfer scenario (Hörcher et al., 2017; Kusakabe et al., 2010), presuming the number of waiting instances is known (Y. Sun & Schonfeld, 2016), or considering train arrival and departure times to be strictly punctual (Zhou & Xu, 2012). Another key challenge is determining the actual train capacity based on travel demand and station characteristics (Mo, Ma, Koutsopoulos, & Zhao, 2020).

Subsequently, some studies began using smart card data for individual travel trip segmentation (Arriagada, Guevara, Munizaga, & Gao, 2024; Luo et al., 2018; L. Sun et al., 2012; F. Zhang et al., 2015). These approaches primarily rely on spatiotemporal constraints to stitch and segment travel trips, enabling large-scale estimation of train choices. However, they are limited by rigid time constraints. Additionally, this method is not well-suited for high-frequency peak-hour scenarios.

Later, researchers began incorporating passenger movement behavior modeling within the urban rail transit system (Tuncel et al., 2023; Zhu et al., 2017a), using maximum likelihood estimation combined with Bayes' theorem to infer train choices. Some studies assume that passenger walking speed follows a log-normal distribution (Zhu et al., 2017b, 2021) and derive initial parameters from survey data. However, these approaches require external data as initial inputs, such as manually collected survey data to establish the initial distribution parameters of passenger walking speeds.

At this stage, some studies apply the EM algorithm directly to process the distribution of walking time (Xiong, Li, Sun, Qin, & Wu, 2022). The core idea is to automatically estimate the distribution parameters of exit walking time using observed travel data, eliminating the need for external initial parameters. Then some researchers extended for single-path individual travel train inference (X. Sun et al., 2024).

Additionally, some studies incorporate train choices estimation in transfer scenarios and validate their models using cross-validation method (C. Li, Xiong, Xiong, Sun, & Qin, 2024). These studies provide valuable methods for data-driven inference of travel trajectories.

In summary, this paper conducts research on inferring complete individual travel trajectories, including selected trains, access/egress/transfer time, and other related factors. This paper proposes a data-driven parameter estimation method for inferring individual trajectories from incomplete data. This method further eliminates dependence on external data or survey data, enabling the automatic extraction of travel location state information. In addition, addressing the issue of previous studies relying on synthetic data for validation, this paper uses real individual travel trajectory data to validate the inference results and confirms the accuracy of our model.

# 3 Problem description

## 3.1 Notation

The necessary symbols and definitions involved in this paper are shown in Table 1.

Table 1. Symbols and Definitions.

| Symbol | Definition |
|---|---|
| $i$ | Travel record index |
| $X_i\ (i \in N)$ | Travel record $i$ |
| $m$ | Travel segment index |
| $M_i$ | Number of segments included in travel record $i$ |
| $X_{i,m}(m \in M_i)$ | Segment $m$ of travel record $i$ |
| $t_i^{in}, t_i^{out}$ | The entry time and exit time of travel record $i$ |
| $t_i^a$ | Access time of travel record $i$ |
| $t_i^e$ | Egress time of travel record $i$ |
| $t_i^{tr}$ | Transfer time of travel record $i$ |
| $t_{i,m}^r$ | Running time of segment $m$ in travel record $i$ |
| $j$ | Train index |
| $Set_{i,m}$ | Train alternative set for the segment $m$ of travel record $i$ |
| $Train_{i,m}^j (j \in Set_{i,m})$ | Train $j$ for segment $m$ of travel record $i$ |
| $DT_{i,m}^j, AT_{i,m}^j$ | The boarding time and alighting time of segment $m$ of travel record $i$ |
| $s_i^o, s_i^{tr}, s_i^d$ | Entry station, transfer station, exit station involved in travel records $i$ |
| $f_e(\cdot)$ | Probability density function of egress time |
| $f_a(\cdot)$ | Probability density function of access time |
| $k_{i,m}$ | Left behind times of segment $m$ in travel record $i$ |
| $P(\text{Train}_{i,m}^j \mid X_i)$ | Probability of traveling on train $j$ on segment $m$ of travel record $i$ |
| $\theta$ | Parameters of egress time distribution |
| $D_o$ | Observable dataset |
| $D_u$ | Unknown dataset |
| $\mu^0, (\sigma^2)^0$ | Initial mean, initial variance of egress time distribution |
| $L(\text{Train} \mid X, \theta^0)$ | Likelihood function for posterior probability |
| $Q(\theta, \theta^0)$ | Log-likelihood function for posterior probability |
| $D_{KL}(\cdot)$ | KL divergence value |
| $\text{Train}_{i,m}^{j*}$ | Train $j^*$ with maximum selection probability for segment $m$ in travel record $i$ |

## 3.2 Problem formulation

AFC data is one of the most important data of urban rail transit system, which mainly records the passenger's entry/exit time and station. Similarly, AVL data is also critical, recording information about the arrival and departure events of urban rail transit trains. However, there is a lack of internal passenger trajectory information, which further leads to the inability to obtain accurate system status of passengers.

In this paper, we utilize AFC data, AVL data, and basic data of the network to design a framework that includes data mining, parameter learning, and probabilistic inference. We aim to achieve a fully data-driven construction of individual passenger travel trajectories within urban rail transit systems. Specifically, this includes individual access/egress/transfer time, train choices, and left behind times. Fig. 1 shows a passenger's complete travel chain.

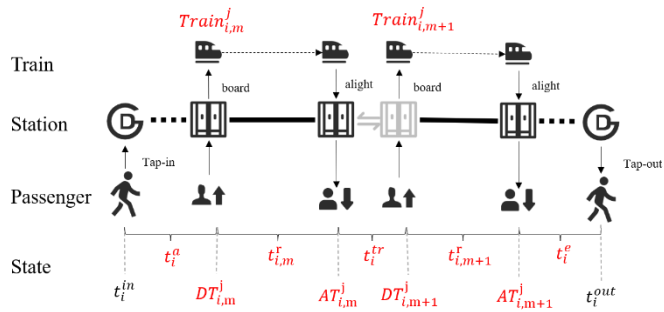

Fig. 1. Schematic diagram of the individual travel chain.

Specifically, for a given OD pair with $N$ travel records, each record is denoted as $X_i, i \in N, X = \{X_1, X_2, \ldots X_i, \ldots X_N,\}$. Each travel record $X_i$ contains $M_i$ segments,

$X_i = \{X_{i,1}, X_{i,2}, \ldots X_{i,m}, \ldots X_{i,M_i}\}, m \in M_i$, where $m$ is the index of segments contained in $X_i$. For example, non-transfer record $X_i = \{X_{i,1}\}$, one transfer record $X_i = \{X_{i,1}, X_{i,2}\}$ and so on. Each record $X_i$ contains passenger elements, time elements, and spatial elements. The reconstruction of individual travel trajectory of urban rail transit is also the process of inferring the elements in $X_i$, denoted as:

$$X_i = \{P_i, T_i, S_i\} \tag{1}$$

For the passenger element $P_i$ is characterized by converting the passenger card number of the AFC data into a passenger index.

For the time element $T_i$, can be divided into entry/exit time, access time and egress time. If the passenger has a transfer, there should also be a transfer time. Among them, the entry/exit time can be obtained directly through the AFC. The access time is the duration after the entry time until boarding the train, and the egress time is the duration from alighting until exit. In addition, the time when passengers board the train after arriving at the platform is the boarding time, which also corresponds to the alighting time. Among the above time elements, the boarding/alighting time can be obtained according to the AVL data after determining the selected train, while the others are unknown.

$$T_i = \{t_i^{in}, t_i^{out}, t_i^{a}, t_i^{e}, t_i^{tr}, DT_{i,m}^{j}, AT_{i,m}^{j}\} \tag{2}$$

For the spatial element $S_i$, the entry/exit stations of the travel are directly available through the AFC data, and if there is a transfer behavior, the transfer stations are also included.

$$S_i = \{s_i^{o}, s_i^{tr}, s_i^{d}\} \tag{3}$$

## 4 Methodology

### 4.1 Model framework

The aim of this paper is to infer individual travel trajectories at the train level from internal data of urban rail transit. This paper proposes a data-driven urban rail travel trajectory inference model, as shown in Fig. 2, supported by real data and probability theory. The methodology's specific process includes establishing train alternative sets based on spatio-temporal constraints, data-driven adaptive trajectory inference, and travel trajectory construction.

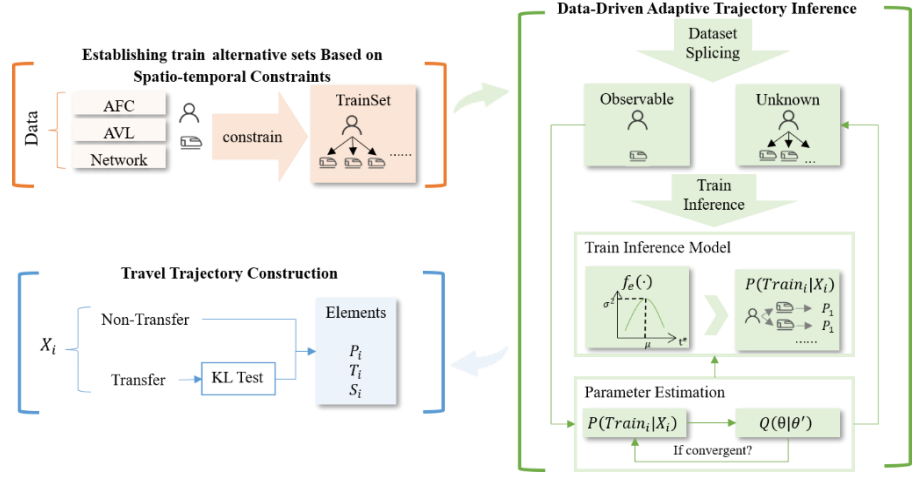

Fig. 2. Model framework.

Briefly, the steps are as follows:

(1) Establishing train alternative sets based on spatio-temporal constraints. By constructing spatio-temporal constraints, passengers are associated with trains to establish a train alternative set for each travel record.

(2) Data-driven adaptive trajectory inference. Establish the train inference model as well as the parameter estimation method to realize automatic trajectory inference. Specifically, it includes dataset slicing, train inference model based on Bayes' theorem, and parameter estimation based on EM algorithm.

(3) Travel trajectory construction. Analyze the travel record elements, and establish the travel trajectory for transfer and non-transfer respectively.

## 4.2 Establishing train alternative sets based on spatio-temporal constraints

Train alternative set $Set_{i,m}$ is the set of possible train choices for each segment $X_{i,m}$ built from AFC data and AVL data through spatio-temporal constraints. The establishment of a train alternative set can reduce the amount of data and thus reduce the computational cost. Fig. 3 illustrates a detailed travel trajectory, where the dashed lines indicate the unboarded trains. And we establish the following spatio-temporal constraints.

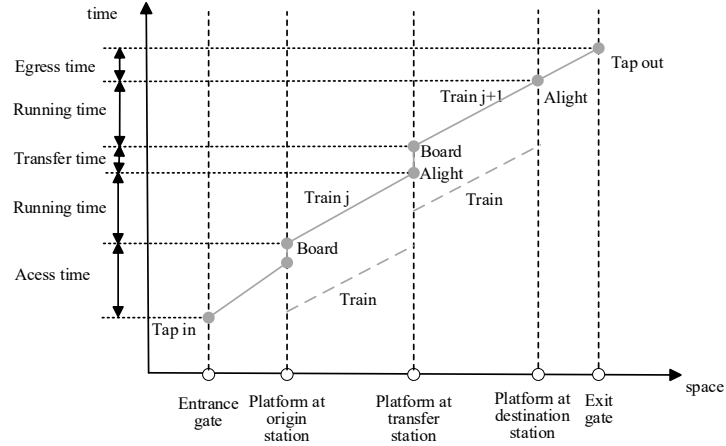

Fig. 3. Spatio-temporal analysis of individual travel trajectory.

(1) Spatial constraints

Passengers travel from the origin station, transfer at the interchange station, and later reach the destination station. Similarly, each segment $X_{i,m}$ requires a train to transport passengers. Fig. 4 illustrates the spatial location of the passenger in relation to the train.

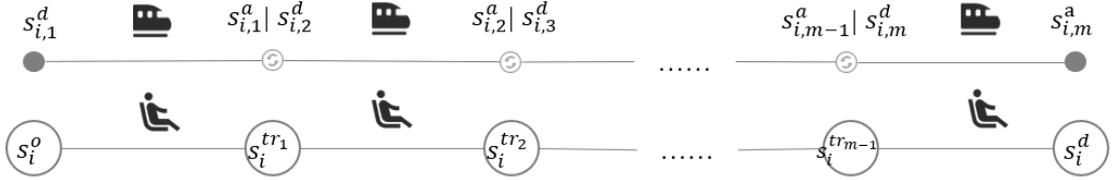

Fig. 4. Spatial location of passengers in relation to the train.

The resulting spatial constraints formed by multiple transfer are as follows:

$$\text{Spatial s.t.} \begin{cases} s_{i,1}^d = s_i^o, s_{i,1}^a = s_i^{tr_1} \\ s_{i,2}^d = s_i^{tr_2}, s_{i,2}^a = s_i^{tr_3} \\ \quad \ldots \ldots \\ s_{i,m}^d = s_i^{tr_{m-1}}, s_{i,m}^a = s_i^d \end{cases} \quad (4)$$

(2) Temporal constraints

The difference between the departure time $DT_{i,m}^j$ and the arrival time $AT_{i,m}^j$ of the train in each segment $X_{i,m}$ is greater than the minimum access time $t_i^{a,min}(t_i^{a,min} \geq 0)$. Similarly, the difference between the exit time $t_i^{out}$ and the arrival time $AT_{i,m}^j$ is greater than the minimum egress time $t_i^{e,min}(t_i^{e,min} \geq 0)$. Thus, the temporal constraints can be formulated as:

$$\text{Temporal s.t.} \begin{cases} DT_{i,m}^j - t_i^{in} \geq t_i^{a,min} \\ t_i^{out} - AT_{i,m}^j \geq t_i^{e,min} \end{cases} \quad (5)$$

In the case of transfer, there should also be a transfer time constraint. The departure time of the train of the latter trip segment $DT_{i,m+1}^j$ should be later than the arrival time of the train of the former trip segment $AT_{i,m+1}^j$.

## 4.3 Data-driven adaptive trajectory inference

### 4.3.1 Dataset slicing

Through the spatio-temporal constraints, establish the train alternative set $Set_{i,m}$ for each segment $X_{i,m}$. If the train in $Set_{i,m}$ is unique, it means the travel record $X_i$ can be solved directly from AFC data and AVL data, that is, the observable data set $D_o$. Conversely, those cannot be determined directly are unknown data sets $D_u$. Fig. 5 presents the process of dataset slicing.

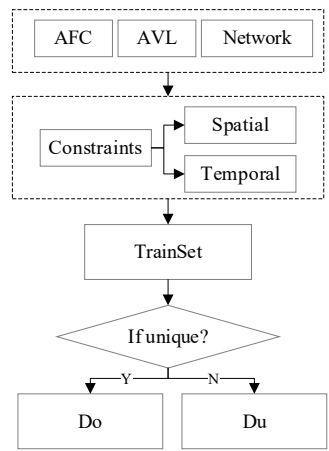

Fig. 5. Flow for dataset slicing.

### 4.3.2 Train Inference Model Based on Bayes' Theorem

After establishing the observable and unknown datasets, a priori samples are generated from the observable data. In this way, the unknown data is trained to avoid dependence on survey data or external data. Thus, a data-driven individual travel trajectory inference is realized.

Inferring the unknown trajectory, the main problem is to determine the selected train. The train alternative set $Set_{i,m}$ gives candidate solutions for each segment $X_{i,m}$, so the problem is converted into a probabilistic problem of solving the optimal selected train in $Set_{i,m}$. Since the priori samples can directly obtain the egress time, if we determine the egress time of the unknown dataset, we can obtain the probability of the train alternative set. Therefore, we focus on solving the distribution of the egress time of the unknown dataset. The main way is by training the distribution parameters of the unknown dataset through the priori samples.

According to Bayes' theorem, the train probability $P(\text{Train}_{i,m}^j \mid X_i)$ means the probability of selecting $\text{Train}_{i,m}^j$ under record $X_i$, and it can be expressed as:

$$P(\text{Train}_{i,m}^j \mid X_i) = \frac{P(X_i \mid \text{Train}_{i,m}^j)P(\text{Train}_{i,m}^j)}{P(X_i)}$$
$$= \frac{P(X_i \mid \text{Train}_{i,m}^j)P(\text{Train}_{i,m}^j)}{\sum_{m=1}^{M_i} \sum_{j'=1}^{\text{Set}_{i,m}} P(X_i \mid \text{Train}_{i,m}^{j'}) P(\text{Train}_{i,m}^{j'})} \qquad (6)$$

Where $P(X_i \mid \text{Train}_{i,m}^j)$ is the possibility of being able to observe $X_i$ under the assumption that the train is $\text{Train}_{i,m}^j$. And the egress time can be calculated as $t_i^{e,j} = t_i^{out} - AT_{i,m=M_i}^j$, let $t_i^{e,j} \sim N(\mu, \sigma^2)$, then $P(X_i \mid \text{Train}_{i,m}^j)$ is denoted as:

$$P(X_i \mid \text{Train}_{i,m}^j) = f_e(t_i^{e,j}; \mu, \sigma^2) \qquad (7)$$

Where $f_e(\cdot)$ is the probability density function of $t_i^{e,j}$.

$P(\text{Train}_{i,m}^j)$ is the prior probability calculated from the observable data $D_o$. With $D_o$ we can obtain the access time $t_i^{a,j}$. It is not difficult to obtain the distribution function of the access time $t_i^{a,j}$ which determines the probability of passengers boarding the train. We regard this probability as a priori probability. Thus, $P(\text{Train}_{i,m}^j)$ is denoted as:

$$P(\text{Train}_{i,m}^j) = f_a(t_i^{a,j}) \qquad (8)$$

Where $t_i^{a,j}$ denotes the access time, which is expressed by the difference between the train departure time $DT_{i,m=1}^j$ and entry time $t_i^{in}$. $f_a(\cdot)$ denotes the probability density function of $t_i^{a,j}$, which obeys the function $t_i^{a,j} \sim N(\mu^a, (\sigma^2)^a)$. Where $\mu^a, (\sigma^2)^a$ are obtained from observable data $D_o$.

Through Eq. (7) and Eq. (8) we can obtain:

$$P(\text{Train}_{i,m}^j \mid X_i) = \frac{P(X_i \mid \text{Train}_{i,m}^j)P(\text{Train}_{i,m}^j)}{P(X_i)}$$
$$= \frac{P(X_i \mid \text{Train}_{i,m}^j)P(\text{Train}_{i,m}^j)}{\sum_{m=1}^{M_i} \sum_{j'=1}^{\text{Set}_{i,m}} P(X_i \mid \text{Train}_{i,m}^{j'}) P(\text{Train}_{i,m}^{j'})}$$
$$= \frac{f_e(t_i^{e,j} \mid \mu, \sigma^2) f_a(t_i^{a,j})}{\sum_{m=1}^{M_i} \sum_{j'=1}^{\text{Set}_{i,m}} f_e(t_i^{e,j'} \mid \mu, \sigma^2) f_a(t_i^{a,j'})} \qquad (9)$$
$$= \frac{f_e(t_i^{out} - AT_{i,m=M_i}^j \mid \mu, \sigma^2) f_a(DT_{i,m=1}^j - t_i^{in})}{\sum_{m=1}^{M_i} \sum_{j'=1}^{\text{Set}_{i,m}} f_e(t_i^{out} - AT_{i,m=M_i}^{j'} \mid \mu, \sigma^2) f_a(DT_{i,m=1}^{j'} - t_i^{in})}$$

Similarly, left behind times $k_{i,m}$, $k_{i,m} = 1,2 \ldots, K_{i,m}$, where $K_{i,m} = len(Set_{i,m}) - 1$, denotes the number of times that a passenger fails to board a train after arriving at the platform. For example, the probability that left behind times $k_{i,m} = 1$ is equal to the probability of a passenger boarding the second train. Thus, we obtain:

$$P(k_{i,m} \mid X_i) = P\left(\text{Train}_{i,m}^{(k_{i,m})+1} \mid X_i\right)$$
$$= \frac{f_e\left(t_i^{out} - AT_{i,m=M_i}^{(k_{i,m})+1} \mid \mu, \sigma^2\right) f_a\left(DT_{i,m=1}^{(k_{i,m})+1} - t_i^{in}\right)}{\sum_{m=1}^{M_i} \sum_{j'=1}^{Set_{i,m}} f_e\left(t_i^{out} - AT_{i,m=M_i}^{j'} \mid \mu, \sigma^2\right) f_a\left(DT_{i,m=1}^{j'} - t_i^{in}\right)} \quad (10)$$

In summary, both $P(\text{Train}_{i,m}^{j} \mid X_i), P(k_{i,m} \mid X_i)$ can be determined by the unknown parameters $\mu, \sigma^2$ in the distribution curve of $f_e(t_i^{e,j}; \mu, \sigma^2)$.

$$\theta = (\mu, \sigma^2) \quad (11)$$

Where $\theta$ denotes the unknown parameter. $\mu$ denotes the mean of egress time distribution. And $\sigma^2$ denotes the variance of egress time distribution.

### 4.3.3 Parameter estimation based on EM algorithm

In previous studies, parameter estimation usually relied on questionnaire surveys. It means fitting parameters through survey data. This method is weakly adaptable and poorly expandable. After constructing the train inference model, we need to obtain the optimal parameters of the egress time distribution. Therefore, we construct a parameter estimation method based on the EM algorithm (Expectation-maximization), which achieves fully data-driven. This method is characterized by iterating the expectation step ("E" step) and maximization step ("M" step).

(1) Initial settings

Fitting the prior distribution of egress time through observable data. Firstly, for the observable dataset $D_o$, the corresponding egress time $t_{i^o}^{e,j}$ is known, from which the distribution curve of $t_{i^o}^{e,j}$ can be solved, expressed as $\theta^0 = (\mu^0, (\sigma^2)^0)$.

$$\mu^0 = \frac{1}{l(D_o)} \sum_{i^o=1}^{l(D_o)} \sum_{j}^{Set_{i^o,m}} t_{i^o}^{e,j} \quad (12)$$

$$(\sigma^2)^0 = \frac{1}{l(D_o) - 1} \sum_{i^o=1}^{l(D_o)} \sum_{j}^{Set_{i^o,m}} \left(t_{i^o}^{e,j} - \mu^0\right)^2 \quad (13)$$

Thus, we obtain the distribution curve parameter $\theta^0$ for the observable data $D_o$. We take $\theta^0$ as the initialization parameter value of the EM algorithm. In this way, we solve the sensitivity problem of parameter initialization.

(2) Expectation calculation

The expected value of train selection is inferred on the basis of the initial setup parameters. Set the hidden variable $\text{Train}_{i,m}^j$, which takes the value of 1 (train is selected) or 0 (train is not selected). So the hidden variable $\text{Train}_{i,m}^j$ a posteriori probability:

$$P(\text{Train}_{i,m}^j \mid X_i, \theta^0) = \frac{f_e\left(t_i^{out} - AT_{i,m=M_i}^j \mid \theta^0\right) f_a\left(DT_{i,m=1}^j - t_i^{in}\right)}{\sum_{m=1}^{M_i} \sum_{j'=1}^{Set_{i,m}} f_e\left(t_i^{out} - AT_{i,m=M_i}^{j'} \mid \theta^0\right) f_a\left(DT_{i,m=1}^{j'} - t_i^{in}\right)} \quad (14)$$

Therefore, the posterior probability of the train is computed given the current parameter $\theta^0$ and used as a weight for the update of the model parameters ("M" step).

(3) Parameter update

The likelihood function of $P(\text{Train}_{i,m}^j \mid X_i, \theta^0)$ can be obtained by Maximum Likelihood Estimation (MLE):

$$L(\text{Train} \mid X, \theta^0) = \prod_i^N \left\{ \frac{f_e\left(t_i^{out} - AT_{i,m=M_i}^j \mid \theta^0\right) f_a\left(DT_{i,m=1}^j - t_i^{in}\right)}{\sum_{m=1}^{M_i} \sum_{j'=1}^{Set_{i,m}} f_e\left(t_i^{out} - AT_{i,m=M_i}^{j'} \mid \theta^0\right) f_a\left(DT_{i,m=1}^{j'} - t_i^{in}\right)} \right\} \quad (15)$$

Through Jensen's inequality, its log-likelihood function $Q(\theta, \theta^0)$ is easily obtained:

$$\begin{aligned} Q(\theta, \theta^0) &= \ell L(\text{Train} \mid X, \theta^0) \\ &= \sum_i^N \ln \frac{f_e\left(t_i^{out} - AT_{i,m=M_i}^j \mid \theta^0\right) f_a\left(DT_{i,m=1}^j - t_i^{in}\right)}{\sum_{m=1}^{M_i} \sum_{j'=1}^{Set_{i,m}} f_e\left(t_i^{out} - AT_{i,m=M_i}^{j'} \mid \theta^0\right) f_a\left(DT_{i,m=1}^{j'} - t_i^{in}\right)} \\ &\geq \sum_i^N \ln \left\{ f_e\left(t_i^{out} - AT_{i,m=M_i}^j \mid \theta^0\right) f_a\left(DT_{i,m=1}^j - t_i^{in}\right) \right\} \\ &\quad - \sum_i^N \ln \left\{ \sum_{m=1}^{M_i} \sum_{j'=1}^{Set_{i,m}} f_e\left(t_i^{out} - AT_{i,m=M_i}^{j'} \mid \theta^0\right) f_a\left(DT_{i,m=1}^{j'} - t_i^{in}\right) \right\} \end{aligned} \quad (16)$$

Update the parameter $\theta'$ by maximizing $Q(\theta, \theta^0)$:

$$\theta' = argmax(Q(\theta, \theta^0)) \quad (17)$$

The updated parameter $\theta'$ is thus obtained. This is used as the input parameter for the "E" step, and the update is iterated. Eventually converge to a maximum likelihood estimate of the parameters. When the algorithm converges, we obtain the parameters of the optimal distribution of egress time. Similarly, we are able to obtain the probability of each train in the train alternative set. Of course, we consider the train with the highest probability as the optimally selected train for the travel record.

## 4.4 Travel Trajectory Construction

The probability of each train in the alternative set is obtained by the data-driven adaptive trajectory inference model. In this section, we mainly construct individual complete travel trajectories through spatio-temporal data associations.

(1) Non-transfer travel trips

For non-transfer trips, the number of travel segments is 1. The maximum selection probability train $\text{Train}_{i,m=1}^{j^*}$ and the number of left behind times $k_{i,m=1}$ are given by Section 4.3. Also using the AVL data, we can obtain its departure time $\text{DT}_{i,m=1}^{j^*}$ at the origin station $s_i^o$ and arrival time $\text{AT}_{i,m=1}^{j^*}$ at the destination station $s_i^d$. Thus, it is easy to obtain:

Travel trip running time $t_{i,m=1}^r$:

$$t_{i,m=1}^r = \text{AT}_{i,m=1}^{j^*} - \text{DT}_{i,m=1}^{j^*} \tag{18}$$

Travel trip time $t_i^e$:

$$t_i^e = t_i^{out} - \text{AT}_{i,m=1}^{j^*} \tag{19}$$

Travel trip time $t_i^a$:

$$t_i^a = \text{DT}_{i,m=1}^{j^*} - t_i^{in} \tag{20}$$

At this point, all inferences have been made for the non-transfer travel trip elements.

(2) Transfer travel trips

For having transfer, Fig. 6 shows the transfer travel trip slicing method. We can slice the travel into segments $X_{i,m}$ by using the transfer station as cut-off point. And apply the adaptive trajectory inference model to solve the train selection for each $X_{i,m}$.

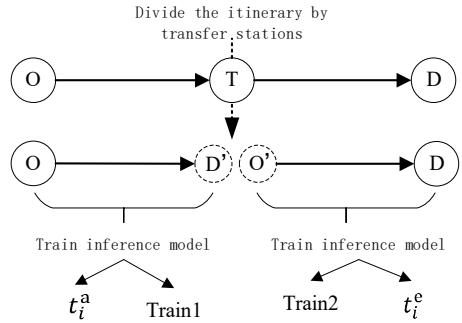

Fig. 6. Schematic diagram of transfer trip segmentation.

Transfer travel trips have a lot of uncertainty. Previous studies usually utilized non-transfer travel trips to estimate transfer travel trips, implying that train inference at the same station and time are considered identical. This approach may introduce biases in the inference of the overall transfer travel trajectories. In other words, this approach may result in the inability to accurately determine the transfer time for the entire travel trip. Therefore, we introduce a Kullback-Leibler Divergence (KL Divergence) constraint to optimize the estimation of transfer time, expressed as:

$$D_{KL}\left(f_{e,X_{i,m}}(\cdot) \parallel f_{e,X_{i,m+1}}(\cdot)\right) = \sum f_{e,X_{i,m}}(\cdot) \log\left(\frac{f_{e,X_{i,m}}(\cdot)}{f_{e,X_{i,m+1}}(\cdot)}\right) \quad (21)$$

Where $f_{e,X_{i,m}}(\cdot)$ denotes the distribution of inferred egress time for the trip segment $X_{i,m}$, and $f_{e,X_{i,m+1}}(\cdot)$ denotes the distribution of inferred egress time for the latter sub-trip segment $X_{i,m+1}$.

Specifically, for $X_i = \{X_{i,1}, X_{i,2}, \ldots X_{i,m}, \ldots X_{i,M_i}\}$, the corresponding probability of each train and the optimal choice of train for the segment $X_{i,m}$ can be inferred by the train inference model. To more accurately infer transfer time, we construct a set of train combinations for transfer travel trips based on the optimal train. Then we construct a matrix of inferred egress time distributions for the train combinations and then calculate the Kullback-Leibler (KL) divergence values for each combination. Through these process, the validation of train combinations is achieved, enabling the determination of the optimal train scheme for the entire transfer travel trips. Fig. 7 illustrates the method of checking train combinations by KL divergence.

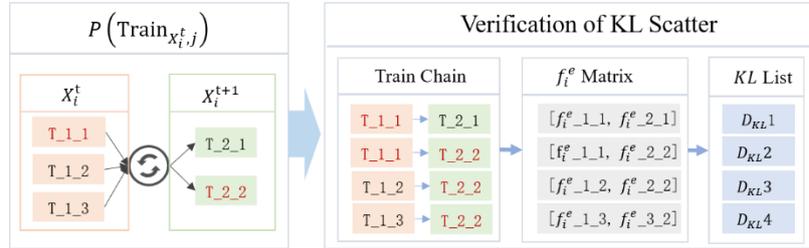

Fig. 7. KL Scatter test inference train combinations.

Then determine whether the combination corresponding to the optimal KL divergence result is consistent with the inferred estimated by EM algorithm. If consistent, then this combination is the overall trajectory combination of the transfer travel trip. Conversely, the KL divergence-minimum combination is used as an input to update the initial parameters of the E-step. And the EM iteration framework is repeated until the optimal travel trajectory combination is obtained.

| KL Divergence Test Inference Train |
|---|
| **Input:** AFC data with transfer, AVL data, basic data of the network |
| 1: Cut $X_i = \{X_{i,1}, X_{i,2}, \ldots X_{i,m}, \ldots X_{i,M_i}\}$ in terms of transfer station |
| 2: **Initialize** $\theta^0 = (\mu^0, (\sigma^2)^0)$ |
| 3: **for** m **in** $M_i$: |

```
 4:     while not converged:
 5:         E-step $P(\text{Train}_{i,m}^j \mid X_i, \theta^0)$
 6:         M-step $\theta' = argmax(Q(\theta, \theta^0))$
 7:         Get $P(\text{Train}_{i,m}^j), f_{e,X_{i,m}}(\cdot)$
 8:     End for
 9:     Get $\{P(\text{Train}_{i,1}^{j^*}), \dots, P(\text{Train}_{i,m}^{j^*})\}, \{f_{e,X_{i,1}}(\cdot), \dots, f_{e,X_{i,m}}(\cdot)\}$
10:     for m in $M_i$-1:
11:         Caculate $D_{KL}\left(f_{e,X_{i,m}}(\cdot) \parallel f_{e,X_{i,m+1}}(\cdot)\right)$
12:     End for
13:     Check $\min(D_{KL}(\cdot)) \Rightarrow (\text{Train}_{i,m}^{j^*}, \text{Train}_{i,m+1}^{j^*})$
14:     If not consistent:
15:         Repeat EM-step
16:     End if
17:     Get $\text{Train}_i^{j^*} = \{\text{Train}_{i,m}^{j^*}, \dots\dots \text{Train}_{i,m}^{j^*}\}$
```

# 5 Case study

## 5.1  Data description

In this paper, CY-BXQ is instantiated using ODs with busy travel during the morning peak period of Beijing urban rail transit. A weekday morning peak (7:00-9:00) travel scenario in 2023 is selected, in which the AFC data totals 1235385 and the AVL data totals 210,112. The case scenario is shown in Fig. 8.

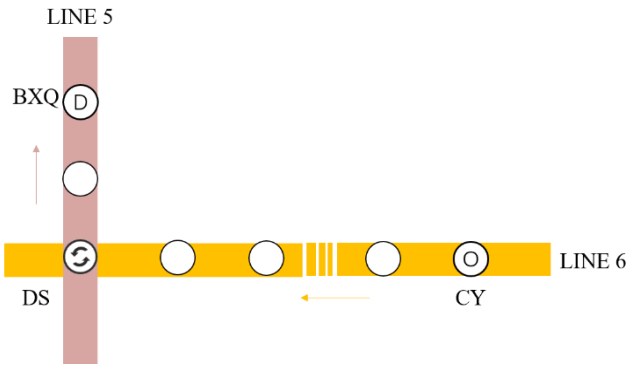

Fig. 8. Case scenario.

To validate the results of the trajectory inference model in this paper, we recruited volunteers to obtain real individual travel trajectory tracking. That is, the individual travel trajectory is reproduced in the real rail transit system. And the actual trajectory of the whole traveling process from entering to exiting the station is recorded. The complete detailed data of the trajectory travel chain is obtained through the experiment. Examples of individual trajectory tracking data are shown in Table 2.

Table 2. Example of individual trajectory tracking data.

| $L_1$ | Boarding train | Boarding time | Boarding station | Alighting time | Alighting station |
|---|---|---|---|---|---|
|  | 1098 | 07:13:48 | CY | 07:37:49 | DS |
| $L_2$ | Boarding train | Boarding time | Boarding station | Alighting time | Alighting station |
|  | 2056 | 07:44:51 | DS | 07:48:04 | BXQ |

## 5.2 Algorithm convergence analysis

The variation of the likelihood function and the estimated mean value of egress time during the parameter fitting process is shown in Fig. 9. During the application of the EM algorithm, the E-step involves solving for the likelihood function, and the M-step focuses on determining the updated parameter values, followed by an update iteration. Therefore, 'old' refers to the solution values based on the pre-update state, while 'new' denotes the solution values after the update.

Fig. 9 demonstrates that the likelihood value gradually increases during the iteration process until it converges, and the mean value of egress time gradually decreases during the iteration process until it converges. In each round of iteration, the model is slowly converged by E-step (expectation step) and M-step (maximization step), indicating that the model parameters gradually converge to the real distribution of the data. At the same time, the parameter update amplitude gradually decreases to 0.001 and tends to stabilize, further verifying the convergence validity of the algorithm.

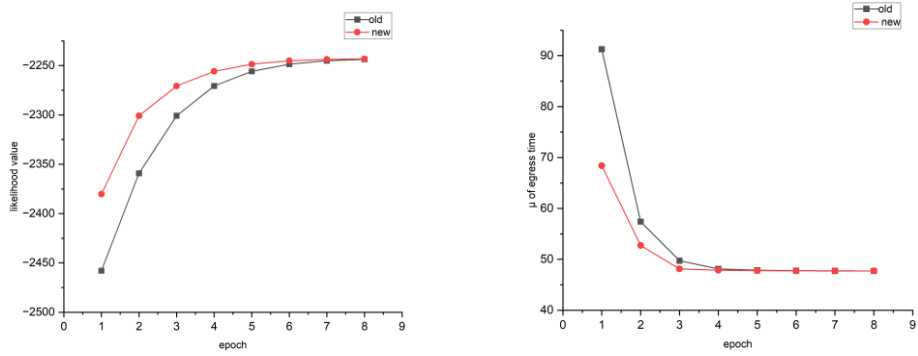

a) Variation in likelihood function.      b) Variation in the mean value of egress time.

Fig. 9. Model iterations during parameter estimation.

## 5.3 Comparative Experimental Analysis

### 5.3.1 Comparison experiment

In this paper, the following methods are mainly selected as comparison experiments:

(1) SSMT: Zhang (F. Zhang et al., 2015) proposed a spatiotemporal segmentation algorithm, Spatiotemporal Segmentation of Metro Trips (SSMT), based on AFC and AVL data. The method finds the nearest neighboring trains matched together with the passenger's exit time as the selected train through a heuristic search method.

(2) PIIM: Zhu (Zhu et al., 2021) proposed a probabilistic method combining Bayes' theorem and maximum likelihood estimation for inferring passenger trip trajectory in urban rail transit, denoted as Passenger Itinerary Inference Model (PIIM). The method assumes that transfer and non-transfer passengers arriving at the platform have the same probability of being left behind. As well as a lognormal distribution with a mean of 1.12 m/s and variance of 0.36 $(m/s)^2$ is used as the egress walking speed distribution based on the observed station characteristics.

### 5.3.2 Evaluation indicators

In this paper, we need to evaluate the performance of the train inference model, which is a multiclassification problem. Therefore, the evaluation indicators in this paper consider both micro-averaging and macro-averaging. Micro-averaging evaluates the performance by calculating the sum of actual and inferred values for all categories.

Macro-averaging evaluates each category and then arithmetically averages the evaluated values of all categories. In other words, micro-averaging focuses on the model's performance in all categories, while macro-averaging focuses on the model's overall performance in different categories. Therefore, the evaluation indicators in this paper are Micro-Precision $P(j)_{micro}$, Micro-recall $R(j)_{micro}$, Micro- F1 score $F1(j)_{micro}$, Macro-Precision $P(j)_{macro}$, Macro-recall $R(j)_{macro}$, Macro- F1 score $F1(j)_{macro}$.

$$\begin{aligned} P(j)_{micro} &= \frac{\sum_{j=1}^{n} C_{j,j}}{\sum_{k=1}^{n}\sum_{j=1}^{n} C_{k,j}} \\ R(j)_{micro} &= \frac{\sum_{j=1}^{n} C_{j,j}}{\sum_{j=1}^{n}\sum_{k=1}^{n} C_{j,k}} \\ F1(j)_{micro} &= 2 \times \frac{P(j)_{micro} \times R(j)_{micro}}{P(j)_{micro} + R(j)_{micro}} \\ P(j)_{macro} &= \frac{1}{n}\sum_{j=1}^{n} P(j)_{micro} \\ R(j)_{macro} &= \frac{1}{n}\sum_{j=1}^{n} R(j)_{micro} \\ F1(j)_{macro} &= \frac{1}{n}\sum_{j=1}^{n} F1(j)_{micro} \end{aligned} \qquad (22)$$

Where $C_{j,j}$ denotes the number of samples where the actual train is $j$ and the inference is also $j$. $\sum_{j=1}^{n} C_{j,j}$ denotes the number of samples in which all inferences are correct. $\sum_{k=1}^{n}\sum_{j=1}^{n} C_{k,j}$ denotes the total sample size of the inferred train for all categories. $\sum_{j=1}^{n}\sum_{k=1}^{n} C_{j,k}$ denotes the total sample size of actual trains for all categories.

### 5.3.3 Results comparison

(1) Comparison of indicators

Taking one transfer as an example, it is cut into two segments according to the transfer station, which are recorded as $L_1, L_2$. Validated by individual travel trajectory tracking data, the evaluation indicators are displayed in Table 3.

Table 3. Comparison of indicators.

| $L_1$ | $P(j)_{micro}$ | $R(j)_{micro}$ | $F1(j)_{micro}$ | $P(j)_{macro}$ | $R(j)_{macro}$ | $F1(j)_{macro}$ |
|---|---|---|---|---|---|---|
| Our's | 0.900 | 0.900 | 0.900 | 0.890 | 0.920 | 0.890 |
| PIIM | 0.800 | 0.800 | 0.800 | 0.714 | 0.690 | 0.698 |

| | SSMT | 0.750 | 0.750 | 0.750 | 0.675 | 0.700 | 0.675 |
|---|---|---|---|---|---|---|---|
| $L_2$ | | $P(j)_{micro}$ | $R(j)_{micro}$ | $F1(j)_{micro}$ | $P(j)_{macro}$ | $R(j)_{macro}$ | $F1(j)_{macro}$ |
| | Our's | 0.950 | 0.950 | 0.950 | 0.882 | 0.882 | 0.882 |
| | PIIM | 0.900 | 0.900 | 0.900 | 0.863 | 0.853 | 0.851 |
| | SSMT | 0.850 | 0.850 | 0.850 | 0.759 | 0.75 | 0.748 |

Factors such as the complexity of operational scenarios and the structural characteristics of stations can affect the train inference model. All the above models perform better for train inference in transfer scenarios. Comparatively speaking, our model has better overall inference performance and higher stability, showing better travel trajectory inference ability.

(2) Matching accuracy

The individual trajectory tracking data is used as a validation set. Calculate the matching accuracy for inferred and actual trains to verify the model. The formula is expressed as:

$$\text{Accuracy} = \frac{\sum_{i=1}^{n} \mathbb{I}(y_i = \hat{y}_i)}{n} \tag{23}$$

Where $y_i$ denotes the actual selected train of sample $i$, $\hat{y}_i$ denotes the inferred selected train of sample $i$, and $n$ denotes the total number of samples in the validation set.

Table 4. Matching accuracy.

| | Our's | PIIM | SSMT |
|---|---|---|---|
| $L_1$ | 90% | 80% | 75% |
| $L_2$ | 95% | 90% | 85% |
| Average | 92.5% | 85% | 80% |

The train inference matching accuracy is shown in Table 4. On the whole, all models perform well. Compared to the other two models, our model achieves superior performance. Overall, our model is well adapted for inferring selected trains for each segment in the transfer scenario.

## 5.4 Trajectory inference results analysis

### 5.4.1 Access/egress time analysis

Taking a single-transfer trip as an example, the train inference model is applied to each segment to identify the train with a maximum probability, which is considered the optimal selected train. Then, it is used as an index to connect with AVL data, thereby constructing the actual individual travel trajectory's access and egress time. Following this, a comparison is made with the actual individual trajectory tracking data, as illustrated in Fig. 10. Overall, the model exhibits minimal estimation error for access and egress time, with the error range confined to [-0.1, 0.1]. It can be concluded that the trajectory inference is relatively close to the actual situation, validating the effectiveness and accuracy of our model.

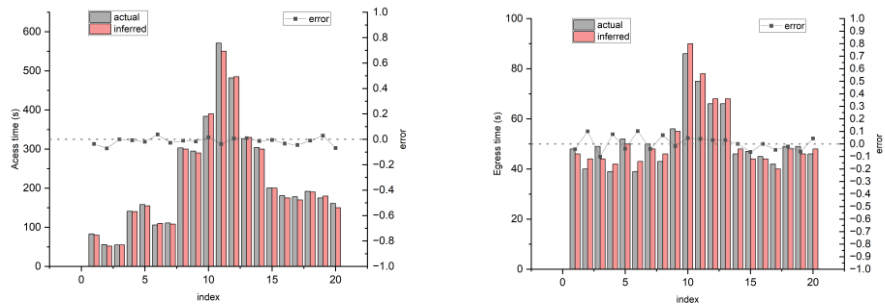

a) Comparison of access time.   b) Comparison of egress time.

Fig. 10. Comparison of actual and inferred access/egress time.

### 5.4.2 Train Inferred analysis

The comparison between the inferred selected trains and the actual selected trains is represented through a confusion matrix, as shown in Fig. 11. Here, the horizontal axis denotes the inferred trains, while the vertical axis represents the actual trains. The percentages indicate the prediction accuracy for each train, with darker shades signifying higher inference accuracy.

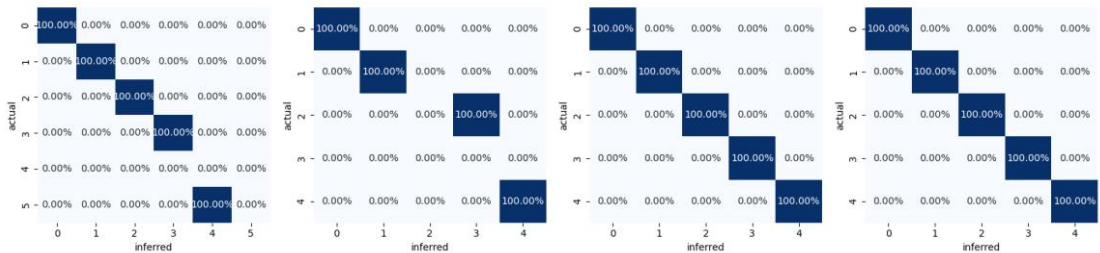

a)   Confusion matrix for $L_1$.

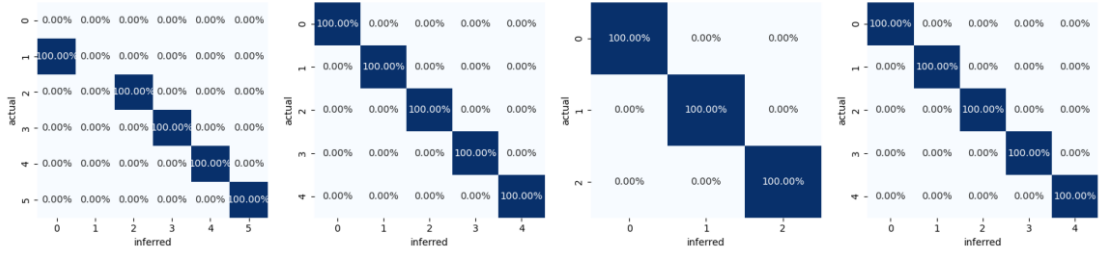

b) Confusion matrix $L_2$.

Fig. 11. Confusion matrix of actual and inferred trains.

Through Fig. 11, it can be found that the inference of the train for each segment is very accurate and closer to the actual trajectory.

### 5.4.3 Transfer time analysis

After obtaining the optimally selected trains for $L_1$ and $L_2$, the inferred transfer time is compared with the actual transfer time as illustrated in Fig. 12. It is observed that the transfer time at this station is predominantly distributed between 200 and 300 seconds. Overall, the model demonstrates an accurate estimation of transfer time and exhibits commendable performance in transfer scenarios.

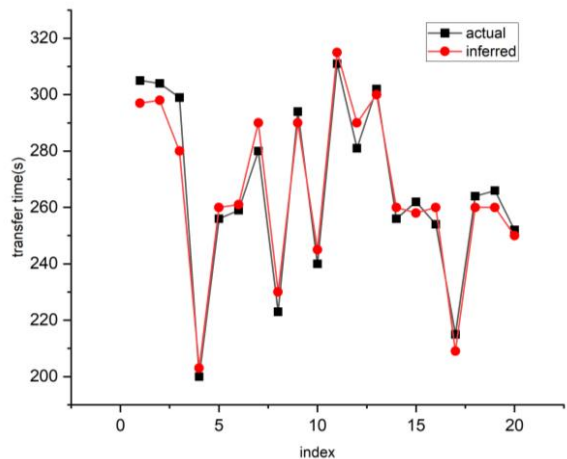

Fig. 12. Comparison of actual and inferred transfer time

### 5.4.4 Left behind analysis

After inferring the optimally selected trains, the number of times passengers are left behind and the corresponding probabilities can be obtained. Fig. 13 presents a comparison between the actual and inferred values of the left-behind times for each passenger under their optimally selected trains. Overall, the inferred left-behind times are quite close to the actual values, demonstrating the model's high accuracy.

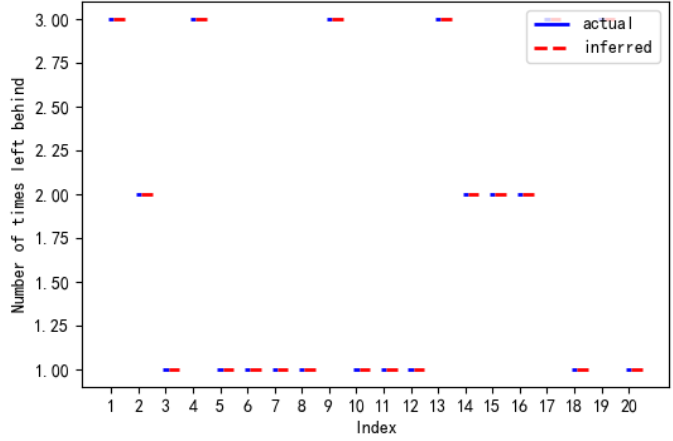

Fig. 13. The number of times a passenger is left behind for actual and inferred.

# 6 Conclusion

In this paper, we present a fully data-driven approach for inferring individual travel trajectories in urban rail systems, applicable to both non-transfer and transfer scenarios. It achieves the deduction of trajectory elements such as selected train, access/egress time, and transfer time. Moreover, this paper presents a method based on real data and probabilistic reasoning, integrating Kullback-Leibler (KL) divergence with the Expectation-Maximization (EM) algorithm to estimate the distribution parameters of unknown egress time. This approach enables data-driven parameter estimation without relying on external or survey data. Additionally, this paper validates the result using real individual travel trajectory tracking data. The analysis of results shows that the inference accuracy exceeds 90%, confirming the model's effectiveness and accuracy.

Future research could expand to include multiple origin-destination (OD) pairs and multi-path studies to realize trajectory inference at the large-scale network level. Of course, we can also improve the model by adding passenger types, station types, or operational strategies as additional model parameters. Future work could also explore the possibility of incorporating real-time samples, based on which real-time individual travel trajectory prediction can be achieved.

**Author Contributions**

**Jie He**：Writing, Software, Methodology, Conceptualization, Formal analysis. Validation. **Yong Qin**: Writing – review & editing, Methodology. **Jianyuan Guo:** Writing – review & editing, Conceptualization. **Xuan Sun**: Data curation, Methodology. **Xuanchuan Zheng**: Funding acquisition, Investigation.

**Acknowledgments**

The authors would like to thank School of Traffic and Transportation, Beijing Jiaotong University and all team members involved in the research work. This work is supported by Beijing Science and Technology Plan Project (Project Number: T22H00010).

# References

1. Arriagada, J., Guevara, C. A., Munizaga, M., & Gao, S. (2024). An experiential learning-based transit route choice model using large-scale smart-card data. *Transportation*, 1-26.
2. Asahara, A., Maruyama, K., Sato, A., & Seto, K. (2011). *Pedestrian-movement prediction based on mixed Markov-chain model.* Paper presented at the Proceedings of the 19th ACM SIGSPATIAL international conference on advances in geographic information systems.
3. Cats, O., & Ferranti, F. (2022). Unravelling individual mobility temporal patterns using longitudinal smart card data. *Research in Transportation Business & Management, 43*, 100816.
4. Chen, X., Cheng, Z., Jin, J. G., Trépanier, M., & Sun, L. (2023). Probabilistic forecasting of bus travel time with a Bayesian Gaussian mixture model. *Transportation Science, 57*(6), 1516-1535.
5. El-Tawab, S., Oram, R., Garcia, M., Johns, C., & Park, B. B. (2016). *Poster: Monitoring transit systems using low cost WiFi technology.* Paper presented at the 2016 IEEE Vehicular Networking Conference (VNC).
6. Gambs, S., Killijian, M.-O., & del Prado Cortez, M. N. (2012). *Next place prediction using mobility markov chains.* Paper presented at the Proceedings of the first workshop on measurement, privacy, and mobility.
7. Gu, J., Jiang, Z., Sun, Y., Zhou, M., Liao, S., & Chen, J. (2021). Spatio-temporal trajectory estimation based on incomplete Wi-Fi probe data in urban rail transit network. *Knowledge-Based Systems, 211*, 106528.
8. Hamdouch, Y., & Lawphongpanich, S. (2008). Schedule-based transit assignment model with travel strategies and capacity constraints. *Transportation Research Part B: Methodological, 42*(7-8), 663-684.
9. Han, G., & Sohn, K. (2016). Activity imputation for trip-chains elicited from smart-card data using a continuous hidden Markov model. *Transportation Research Part B: Methodological, 83*, 121-135.
10. Hörcher, D., Graham, D. J., & Anderson, R. J. (2017). Crowding cost estimation with large scale smart card and vehicle location data. *Transportation Research Part B: Methodological, 95*, 105-125.
11. Hsu, Y.-W., Wang, T.-Y., & Perng, J.-W. (2020). Passenger flow counting in buses based on deep learning using surveillance video. *Optik, 202*, 163675.
12. Kapatsila, B., Bahamonde-Birke, F. J., van Lierop, D., & Grisé, E. (2025). The effect of crowding level information provision on the revealed route choice of transit riders. *Transportation*, 1-26.

13. Kusakabe, T., Iryo, T., & Asakura, Y. (2010). Estimation method for railway passengers' train choice behavior with smart card transaction data. *Transportation, 37*, 731-749.
14. Lesani, A., & Miranda-Moreno, L. (2018). Development and testing of a real-time WiFi-bluetooth system for pedestrian network monitoring, classification, and data extrapolation. *IEEE Transactions on Intelligent Transportation Systems, 20*(4), 1484-1496.
15. Li, B., Yao, E., Yamamoto, T., Huan, N., & Liu, S. (2020). Passenger travel behavior analysis under unplanned metro service disruption: Using stated preference data in Guangzhou, China. *Journal of Transportation Engineering, Part A: Systems, 146*(2), 04019069.
16. Li, C., Xiong, S., Xiong, H., Sun, X., & Qin, Y. (2024). Logistic model for pattern inference of subway passenger flows based on fare collection and vehicle location data. *Applied Mathematical Modelling, 130*, 472-495.
17. Liu, M., Ning, J., Du, Y., Cao, J., Zhang, D., Wang, J., & Chen, M. (2020). Modelling the evolution trajectory of COVID-19 in Wuhan, China: experience and suggestions. *Public health, 183*, 76-80.
18. London, M. o. (2024). *Travel in London 2024 Annual overview* Retrieved from https://content.tfl.gov.uk/travel-in-london-2024-annual-overview-acc.pdf
19. Luo, D., Bonnetain, L., Cats, O., & van Lint, H. (2018). Constructing spatiotemporal load profiles of transit vehicles with multiple data sources. *Transportation Research Record, 2672*(8), 175-186.
20. Ma, X., Wu, Y.-J., Wang, Y., Chen, F., & Liu, J. (2013). Mining smart card data for transit riders' travel patterns. *Transportation Research Part C: Emerging Technologies, 36*, 1-12.
21. Mo, B., Ma, Z., Koutsopoulos, H. N., & Zhao, J. (2020). Capacity-constrained network performance model for urban rail systems. *Transportation Research Record, 2674*(5), 59-69.
22. Mo, B., Zhao, Z., Koutsopoulos, H. N., & Zhao, J. (2021). Individual mobility prediction in mass transit systems using smart card data: An interpretable activity-based hidden Markov approach. *IEEE Transactions on Intelligent Transportation Systems, 23*(8), 12014-12026.
23. Myrvoll, T. A., Håkegård, J. E., Matsui, T., & Septier, F. (2017). *Counting public transport passenger using WiFi signatures of mobile devices.* Paper presented at the 2017 IEEE 20th International Conference on Intelligent Transportation Systems (ITSC).
24. Nuzzolo, A., Crisalli, U., & Rosati, L. (2012). A schedule-based assignment model with explicit capacity constraints for congested transit networks. *Transportation Research Part C: Emerging Technologies, 20*(1), 16-33.
25. Raveau, S., Muñoz, J. C., & De Grange, L. (2011). A topological route choice model for metro. *Transportation Research Part A: Policy and Practice, 45*(2), 138-147.
26. Shi, Z., Pan, W., He, M., & Liu, Y. (2023). Understanding passenger route choice behavior under the influence of detailed route information based on smart card data. *Transportation*, 1-25.
27. Si, B., Zhong, M., Liu, J., Gao, Z., & Wu, J. (2013). Development of a transfer-cost-based logit assignment model for the Beijing rail transit network using automated fare collection data. *Journal of Advanced Transportation, 47*(3), 297-318.
28. Su, G., Si, B., Zhao, F., & Li, H. (2022). Data-Driven Method for Passenger Path Choice Inference in Congested Subway Network. *Complexity, 2022*(1), 5451017.
29. Sun, L., Lee, D.-H., Erath, A., & Huang, X. (2012). *Using smart card data to extract passenger's spatio-temporal density and train's trajectory of MRT system.* Paper presented at the Proceedings of the ACM SIGKDD International Workshop on Urban Computing, Beijing, China.

30. Sun, L., Lu, Y., Jin, J. G., Lee, D.-H., & Axhausen, K. W. (2015). An integrated Bayesian approach for passenger flow assignment in metro networks. *Transportation Research Part C: Emerging Technologies, 52*, 116-131.

31. Sun, X., Guo, J., Qin, Y., Zheng, X., Xiong, S., He, J., . . . Jia, L. (2024). A Spatiotemporal Probabilistic Graphical Model Based on Adaptive Expectation-Maximization Attention for Individual Trajectory Reconstruction Considering Incomplete Observations. *Entropy, 26*(5), 388.

32. Sun, Y., & Schonfeld, P. M. (2016). Schedule-based rail transit path-choice estimation using automatic fare collection data. *Journal of Transportation Engineering, 142*(1), 04015037.

33. Tian, Y., Zhu, W., & Song, F. (2024). Route choice modelling for an urban rail transit network: past, recent progress and future prospects. *European Transport Research Review, 16*(1), 52.

34. Transport, M. o. (2024). *National Urban Passenger Volume from January to November 2024*. Retrieved from https://xxgk.mot.gov.cn/2020/jigou/zhghs/202412/t20241224_4161562.html

35. Tuncel, K. S., Koutsopoulos, H. N., & Ma, Z. (2023). An Unsupervised Learning Approach for Robust Denied Boarding Probability Estimation Using Smart Card and Operation Data in Urban Railways. *IEEE Intelligent Transportation Systems Magazine*.

36. Wang, J., Wu, N., Zhao, W. X., Peng, F., & Lin, X. (2019). *Empowering A\* search algorithms with neural networks for personalized route recommendation.* Paper presented at the Proceedings of the 25th ACM SIGKDD international conference on knowledge discovery & data mining.

37. Weng, X., Liu, Y., Song, H., Yao, S., & Zhang, P. (2018). Mining urban passengers' travel patterns from incomplete data with use cases. *Computer Networks, 134*, 116-126.

38. Wu, J., Qu, Y., Sun, H., Yin, H., Yan, X., & Zhao, J. (2019). Data-driven model for passenger route choice in urban metro network. *Physica A: Statistical Mechanics and its Applications, 524*, 787-798.

39. Xiong, S., Li, C., Sun, X., Qin, Y., & Wu, C. F. J. (2022). Statistical estimation in passenger-to-train assignment models based on automated data. *Applied Stochastic Models in Business and Industry, 38*(2), 287-307.

40. Yin, H., Wu, J., Liu, Z., Yang, X., Qu, Y., & Sun, H. (2019). Optimizing the release of passenger flow guidance information in urban rail transit network via agent-based simulation. *Applied Mathematical Modelling, 72*, 337-355.

41. Zhang, F., Zhao, J., Tian, C., Xu, C., Liu, X., & Rao, L. (2015). Spatiotemporal segmentation of metro trips using smart card data. *IEEE Transactions on Vehicular Technology, 65*(3), 1137-1149.

42. Zhang, Y.-S., & Yao, E.-J. (2015). Splitting Travel Time Based on AFC Data: Estimating Walking, Waiting, Transfer, and In-Vehicle Travel Times in Metro System. *Discrete Dynamics in Nature and Society, 2015*(1), 539756.

43. Zhao, J., Qu, Q., Zhang, F., Xu, C., & Liu, S. (2017). Spatio-temporal analysis of passenger travel patterns in massive smart card data. *IEEE Transactions on Intelligent Transportation Systems, 18*(11), 3135-3146.

44. Zhao, J., Zhang, L., Ye, K., Ye, J., Zhang, J., Zhang, F., & Xu, C. (2022). GLTC: A metro passenger identification method across AFC data and sparse wifi data. *IEEE Transactions on Intelligent*

*Transportation Systems, 23*(10), 18337-18351.
45. Zhou, F., & Xu, R.-h. (2012). Model of passenger flow assignment for urban rail transit based on entry and exit time constraints. *Transportation Research Record, 2284*(1), 57-61.
46. Zhu, Y., Koutsopoulos, H. N., & Wilson, N. H. (2017a). Inferring left behind passengers in congested metro systems from automated data. *Transportation research procedia, 23*, 362-379.
47. Zhu, Y., Koutsopoulos, H. N., & Wilson, N. H. (2017b). A probabilistic passenger-to-train assignment model based on automated data. *Transportation Research Part B: Methodological, 104*, 522-542.
48. Zhu, Y., Koutsopoulos, H. N., & Wilson, N. H. (2021). Passenger itinerary inference model for congested urban rail networks. *Transportation Research Part C: Emerging Technologies, 123*, 102896.